\pdfoutput=1
\documentclass[11pt]{article}

\usepackage[final]{acl}

\usepackage{times}
\usepackage{latexsym}

\usepackage[T1]{fontenc}
\usepackage[utf8]{inputenc}

\usepackage{microtype}

\usepackage{inconsolata}

\usepackage{graphicx}

\usepackage{makecell}
\usepackage{bbding}
\usepackage{multirow}
\usepackage{pgfplots}
\usepackage{svg}
\usepackage{amsmath} 
\usepackage{amssymb}
\usepackage{url}
\usepackage{tablefootnote}
\usepackage{graphicx}
\usepackage{cuted}
\usepackage{tabularx}
\usepackage{xcolor}
\usepackage{pifont}

\usepackage{adjustbox}

\usepackage{listings}
\newcommand{\benchmarkname}{FAME}
\newcommand{\methodname}{SKEME}
\newcommand{\scorename}{SURE}
\definecolor{deepgreen}{RGB}{6,153,6} 
\definecolor{deepred}{RGB}{254,34,35}    
\title{\benchmarkname: Towards Factual Multi-Task Model Editing}
\author{
 \textbf{Li Zeng\textsuperscript{1\footnotemark[1]}},
 \textbf{Yingyu Shan\textsuperscript{1\footnotemark[1]}},
 \textbf{Zeming Liu\textsuperscript{2\footnotemark[1]}},
 \textbf{Jiashu Yao\textsuperscript{1}},
 \textbf{Yuhang Guo\textsuperscript{1\footnotemark[2]}}
\\
 \textsuperscript{1}School of Computer Science and Technology, Beijing Institute of Technology, Beijing, China
\\
 \textsuperscript{2}School of Computer Science and Engineering, Beihang University, Beijing, China
\\
 \href{zengli@bit.edu.cn}{\textcolor{black}{\{zengli}}, 
\href{shanyingyu@bit.edu.cn}{\textcolor{black}{shanyingyu}}, 
\href{yaojiashu@bit.edu.cn}{\textcolor{black}{yaojiashu}}, 
\href{guoyuhang@bit.edu.cn}{\textcolor{black}{guoyuhang\}}}@bit.edu.cn
\href{zmliu@buaa.edu.cn}{\textcolor{black}{zmliu}}@buaa.edu.cn
}
\begin{document}
\maketitle
\begin{abstract}
\renewcommand{\thefootnote}{\fnsymbol{footnote}} 
\footnotetext[1]{Equal contribution}
\footnotetext[2]{Corresponding author: \href{guoyuhang@bit.edu.cn}{guoyuhang@bit.edu.cn}}
\renewcommand{\thefootnote}{\arabic{footnote}}

Large language models (LLMs) embed extensive knowledge and utilize it to perform exceptionally well across various tasks. Nevertheless, outdated knowledge or factual errors within LLMs can lead to misleading or incorrect responses, causing significant issues in practical applications.
To rectify the fatal flaw without the necessity for costly model retraining, various model editing approaches have been proposed to correct inaccurate knowledge within LLMs in a cost-efficient way.
To evaluate these model editing methods, previous work introduced a series of datasets. 
However, most of the previous datasets only contain fabricated data in a single format, which diverges from real-world model editing scenarios, raising doubts about their usability in practice.
To facilitate the application of model editing in real-world scenarios, we propose the challenge of practicality.
To resolve such challenges and effectively enhance the capabilities of LLMs, we present \benchmarkname, an factual, comprehensive, and multi-task dataset, which is designed to enhance the practicality of model editing. We then propose \methodname, a model editing method that uses a novel caching mechanism to ensure synchronization with the real world.
The experiments demonstrate that \methodname~performs excellently across various tasks and scenarios, confirming its practicality.
\footnote{Dataset and codes are publicly available at \url{https://github.com/BITHLP/FAME}}

\end{abstract}

\section{Introduction}
\label{sec:introduction}
Large language models (LLMs) have achieved remarkable capabilities across various domains and are extensively utilized in practical applications \citep{touvron2023Llama,touvron2023Llama2,OpenaiGPT4,geva2020transformer,geva2022transformer}. The extensive utilization of LLMs makes it essential for them to provide precise information. However, LLMs may still provide erroneous information due to incorrect, outdated knowledge stored within the model \citep{decaoetal2021editing, agarwal2022temporal}. Such erroneous information can have significant repercussions within critical domains like medical diagnostics and legal consultations, underscoring the importance of rectifying errors in language models.
To avoid costly retraining and to efficiently correct the outputs of LLMs, model editing has been proposed \citep{mitchell2022memory,Sinitsin2020Editable,decaoetal2021editing}.

\begin{figure}[t]
    \centering
    \includegraphics[width=\linewidth]{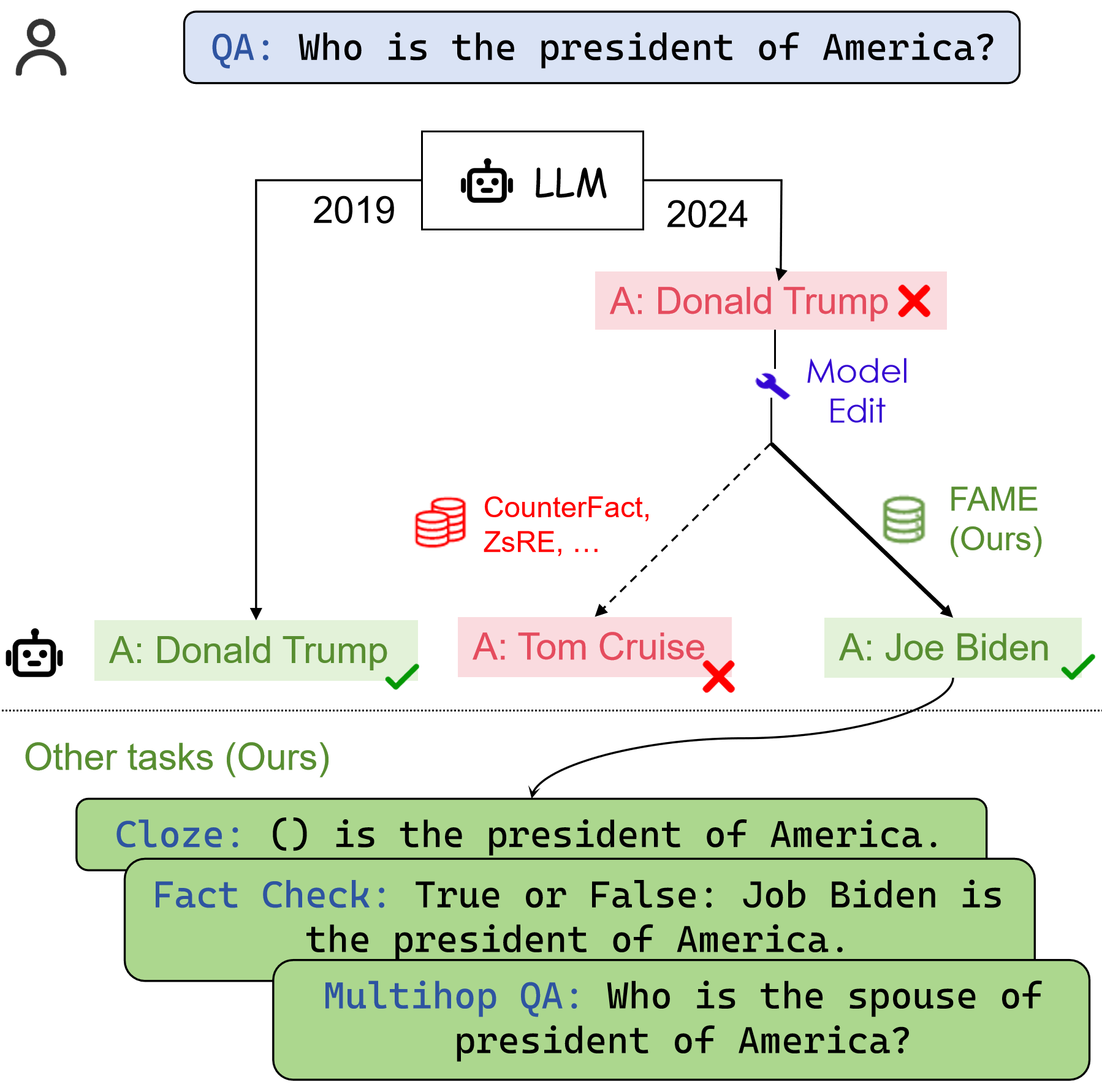}
    \caption{\label{figure:example}An example of \benchmarkname. LLMs may develop factual inaccuracies over time, which can be corrected through model editing. While previous datasets employed fabricated data, \benchmarkname~utilizes real-world data to improve the performance of LLMs in practical usage.
    }
\end{figure}

To evaluate model editing methods, previous works have introduced a series of datasets \citep{decaoetal2021editing,meng2022locating,zhong2023MQuAKE}. 
Almost all of these datasets set the target as incorrect answers, which affects the model's practical performance and contradicts the original purpose of model editing. As shown in Figure \ref{figure:example}, when the user asks "Who is the President of America?", LLMs produce incorrect output due to outdated knowledge. Previous datasets \citep{levy2017zero,meng2022locating,gupta2023editing} modified them to other wrong targets (for example, Tom Cruise).
Moreover, these datasets are all composed of data in a single format with a single task like QA \citep{levy2017zero} or sentence completion \citep{meng2022locating}, which leads to a disparity between experiments and practical applications.

To promote the capability of model editing in practical applications, we introduce a novel criterion: \textbf{Practicality}. 
Practicality refers to the capacity of data and methods to be functional in real-world applications. This entails that data should be factual, diverse, and of high quality, and methods should be efficient and general across tasks. These requirements collectively ensure the effectiveness and applicability of model editing in practical applications.

To address the practical shortcomings in the previous benchmark, which mainly include incorrect knowledge and limited task formats, we introduce \benchmarkname, a factual, extensive model editing benchmark with practicality. \benchmarkname~comprises 128k real data items, including various tasks with single-hop and multi-hop questions. In response to incorrect knowledge, we extract factual data items from Wikidata \citep{vrandevcic2014Wikidata} and  DBpedia \citep{auer2007DBpedia} and employ multiple rounds of manual verification to ensure the accuracy of data items. 
To prevent limited task formats, we incorporate tasks from existing datasets like QA \citep{levy2017zero}, fact-check \citep{schuster2021get}, multi-hop QA \citep{zhong2023MQuAKE}, and introduce new tasks such as cloze and dialogue, making our evaluation more comprehensive.
\benchmarkname~ enhances the model's ability to solve real-world problems and cross-domain issues and enables a complete evaluation of the effectiveness of model editing.

Aimed at tackling the deficiency in the practicality of previous methods, we introduce \methodname. \methodname~ utilizes a novel caching mechanism and receives information from diverse sources to ensure synchronization with the real world, allowing for the application of \methodname~in real-world scenarios. 
The caching mechanism tackles the challenges posed by large-scale data and diverse tasks.

To evaluate the practicality of model editing methods, 
we first introduce a new metric, \scorename, which considers both accuracy and side effects and is adaptable to scenarios. We then evaluate whether each method can meet the basic requirements of model editing (Section \ref{sec:main_exp}). Subsequently, We discuss the performance of model editing methods in real-world scenarios  (Section \ref{sec:Analysis}). The conclusion indicates that previous methods either exhibit side effects or struggle to handle complex scenarios, while only \methodname~consistently outperformed others across all experimental conditions, demonstrating the superiority of \methodname~in real-world scenarios.

The main contributions of this paper are as follows:
    \begin{itemize}
        \item We first introduce the practicality requirement for model editing, which necessitates data and methods to exhibit effective performance in real-world applications.

         \item To support the practicality requirement of model editing, we create \benchmarkname, a novel benchmark that utilizes real-world knowledge and common diverse tasks to simulate practical applications.

    \item To meet the practicality requirement of the model editing 
     method, we propose a method called \methodname, which is the first to introduce a novel caching mechanism for efficient storage, retrieval, and update of constantly evolving real-world facts. Experiments demonstrate that \methodname~is more effective in real-world scenarios.
    \end{itemize}

\section{Related work}

    \subsection{Model Editing Datasets}

    Model editing datasets serve the purpose of verifying the effectiveness of methods and enhancing the capability of LLMs. Nevertheless, current datasets fall short of directly enhancing the capability of LLMs. The majority of datasets comprise constructed fake data \citep{levy2017zero,meng2022locating,gupta2023editing}, primarily serving to validate effectiveness rather than directly contribute to the enhancement of LLMs' capabilities.
    MQuAKE-T \citep{zhong2023MQuAKE} utilizes modifications in Wikidata, which has the potential to directly enhance LLMs' practical performance. 
    However, due to the limited combinations of relations (see Figure \ref{fig:cmp_multi_qa}~for statistics), its direct utility in improving the performance of LLMs is limited, thereby primarily serving to validate effectiveness.
    In contrast to prior works, our benchmark sets itself apart by featuring a substantial repository of factual data and integrating multiple diverse tasks. As a result, it exhibits a heightened level of practical applicability.

    \subsection{Model Editing Methods}

    Previous works have introduced various model-editing methods, both parameter modification and parameter preservation approaches \citep{yao2023editing}. The former category includes the locate-then-edit method \citep{meng2022locating,meng2022mass}  and meta-learning-based methods \citep{decaoetal2021editing,mitchell2021fast}. The latter category involves adding additional parameters to the model \citep{huang2023transformer} and employing vector databases for knowledge storage and retrieval \citep{mitchell2022memory,zhong2023MQuAKE,zheng2023can,cheng2023decouple,madaan2022memory}. Building on this foundation, there are also methods such as reflection \citep{wang2024memoryllm},  optimization of searches for multi-hop questions \citep{shi2024retrievalenhanced}, and the use of post-processing \citep{song2024knowledge} to improve retrieval strategies. Diverging from the previously discussed methods, our method involves a novel caching mechanism, allowing for the application in real-world scenarios.

\section{Problem Definition}
\label{sec:problemsDefinition}   

The objective of model editing is to modify the knowledge contained in a model, allowing the model to engage in reasoning processes based on the edited knowledge, while not affecting the output related to the unedited knowledge. Based on previous work \citep{wang2023knowledge,yao2023editing}, we define model editing to express the goal as follows.

    An input-output pair is defined as $(x, y)$, and a model is represented by a function \(f: X \rightarrow Y\), where $X$ represents the input set and $Y$ represents the output set. Let \(I(x, y)\) denotes the set of descriptions semantically equivalent to \((x, y)\), and \(EX(x, y)\) be the set of input-output pairs that the model can possess with \(I(x, y)\) as prior knowledge. Then, let \(O(x, y)\) represent the portion outside \(I(x, y)\) and \(EX(x, y)\). Appendix \ref{sec:exampleOfDefination} provides an example of the definition.

Formally,  let \((\text{subject, relation, object})\) be a factual triple, denoted as \((s, r, o)\). Consider an input-output pair as \((x, y)\), where \(x\) is effectively a combination of \(s\) and \(r\). A model is represented by a function \(f: X \rightarrow Y\), where $X$ represents the input set and $Y$ represents the output set.

    We use the prime notation to denote semantically equivalent elements. Specifically, for any \(t\) in the set \(\{s, r, o, x, y\}\), let \(t'\) be any element that is semantically equivalent to \(t\), and let \(T'\) be the set of all such \(t'\). Notice that \(t \in T'\).
    Then, we can define \(I(x, y)\) as    
\begin{equation}
\setlength{\abovedisplayskip}{2pt} 
\setlength{\belowdisplayskip}{2pt}
    I(x, y) = \{(x', y') | x' \in X' \text{ and } y' \in Y'\}.
\end{equation}
    To define \(EX(x, y)\), let's represent a fact triple as \(tr(s, r, o)\), abbreviated as \( tr \), and \(S\) is the set of all fact triples. Also, define the multiplication operation $*$ for two sets of fact triples A and B as the join operation: 
\begin{equation}
\setlength{\abovedisplayskip}{2pt} 
\setlength{\belowdisplayskip}{2pt}
    A*B = A\mathop{\Join}\limits_{o=s} B
\end{equation}
Then, define 
\begin{equation}
\setlength{\abovedisplayskip}{2pt} 
\setlength{\belowdisplayskip}{2pt}
    N_0(tr) = \{(s', r', o') \mid s' \in S', r' \in R', o' \in O'\}
\end{equation} 
and 
\begin{equation}
\setlength{\abovedisplayskip}{2pt} 
\setlength{\belowdisplayskip}{2pt}
    N_i(tr) = N_{i-1}(tr) * S  \quad   \quad  (i \geq 1)\\
\end{equation}
Ultimately, we define \(EX(tr)\) as
\begin{equation}
\setlength{\abovedisplayskip}{2pt} 
\setlength{\belowdisplayskip}{2pt}
    EX(tr) =\bigcup\limits_{i=0}\limits^{\infty} N_i
\end{equation}
By transforming \( s \) and \( r \) into \( x \), and \( o \) into \( y \), we derive \(EX(x, y)\).

After defining \(I(x, y)\) and \(EX(x, y)\), we can define \(O(x, y)\) as 
\begin{equation}    
\setlength{\abovedisplayskip}{2pt} 
\setlength{\belowdisplayskip}{2pt}
    O(x, y) = \complement_S(I(x, y) \cup EX(x, y))
\end{equation}
where \(\complement_S\) represents the complement within the set \(S\).

The definition of model editing can be summarized as follows: $(x_f, y_f)$ denotes the fact that is being edited, while $(x_e, y_e)$ represents the input and output.
\begin{equation}
\fontsize{10}{\baselineskip}\selectfont
\setlength{\abovedisplayskip}{0pt} 
\setlength{\belowdisplayskip}{3pt}
\label{eq:prev_ternary}
  f'(x_e) =\! \left\{
         \begin{array}{lr}
         \!\! y_f &  (x_e,y_e)\in I(x_f, y_f) \\
         \!\! f(x_e) \mid (x_f, y_f) &  (x_e,y_e)\in EX(x_f, y_f) \\
         \!\! f(x_e)  &  (x_e,y_e)\in O(x_f, y_f) \\
         \end{array}
         \right.
        \nonumber 
\end{equation}

\begin{table*}[!ht]
\centering
    \tabcolsep=0.15cm
    \begin{tabular}{cccccccccccc} 
\hline

      \multirow{2}*{\textbf{Name}}& \multirow{2}*{\textbf{isC.}}&\multicolumn{6}{c}{\textbf{Tasks}}&\multirow{2}*{\textbf{Total}}& \multirow{2}*{\textbf{Re.}}& \multirow{2}*{\textbf{Source}}& \multirow{2}*{\textbf{Hop}} \\
      
		\cline{3-8}
      ~& ~& \textbf{Cho.}&\textbf{FC.}& \textbf{Clo.} &\textbf{Dia.}& \textbf{Com.}& \textbf{QA}&~ &~ &~ &~\\
\hline
      
      \textbf{\textsc{ZsRE} } &  \color{deepred}\XSolidBrush &\color{deepred}\XSolidBrush &\color{deepred}\XSolidBrush &\color{deepred}\XSolidBrush &\color{deepred}\XSolidBrush &\color{deepred}\XSolidBrush &\color{deepgreen}\CheckmarkBold &270K &  120 & WD. & Si.\\
      \textbf{\textsc{CounterFact}} &  \color{deepred}\XSolidBrush & \color{deepred}\XSolidBrush &\color{deepred}\XSolidBrush &\color{deepred}\XSolidBrush &\color{deepred}\XSolidBrush &\color{deepgreen}\CheckmarkBold   & \color{deepred}\XSolidBrush &2.2K & 24&  WD. & Si. \\
      \textbf{\textsc{MQuAKE-cf} } &  \color{deepred}\XSolidBrush & \color{deepred}\XSolidBrush &\color{deepred}\XSolidBrush &\color{deepred}\XSolidBrush &\color{deepred}\XSolidBrush &\color{deepred}\XSolidBrush  &\color{deepgreen}\CheckmarkBold & 9.2K  & 37&  WD. & Mu. \\
      \textbf{\textsc{\mbox{MQuAKE-t}} }&  \color{deepgreen}\CheckmarkBold&\color{deepred}\XSolidBrush &\color{deepred}\XSolidBrush &\color{deepred}\XSolidBrush &\color{deepred}\XSolidBrush &\color{deepred}\XSolidBrush &\color{deepgreen}\CheckmarkBold&1.8K  & 6 & WD. & Mu. \\
      
\hline
      \textbf{\textsc{\benchmarkname~(Ours)} } & \color{deepgreen}\CheckmarkBold  & \color{deepgreen}\CheckmarkBold & \color{deepgreen}\CheckmarkBold & \color{deepgreen}\CheckmarkBold & \color{deepgreen}\CheckmarkBold& \color{deepgreen}\CheckmarkBold&\color{deepgreen}\CheckmarkBold& 128K  & 86 &
      
     WD. \& DB.   & Si. \& Mu. \\
\hline
\end{tabular}
 \caption{\label{dataset_cmp}
Comparison between \benchmarkname~ to other model edit datasets. 
"isC." stands for isCorrect, which means if the edit target is the real fact.
"Cho.", "FC.", "Clo.", "Dia.", "Com.", "Re", "WD.", "DB.", "Si.", "Mu." stands for choose, fact-check, cloze, dialogue, completion, Relations, Wikidata, DBpeida, single-hop data, and multi-hop data, respectively.
}
\end{table*}

\section{\benchmarkname: A Practical Model Editing Benchmark}

\benchmarkname~(\textbf{FA}ctual \textbf{M}ulti-task model \textbf{E}diting) is a benchmark comprising 128k factual data items. We utilize these data items to construct both single-hop and multi-hop questions. For single-hop questions, we include six forms: QA, sentence completion, cloze test, multiple-choice questions, fact check, and locality test. For multi-hop questions, we include multi-hop questions and dialogues. The previous work introduced QA, sentence completion, fact check, and multi-hop questions \citep{wang2023knowledge}, while we propose the remaining tasks. We believe that combining these tasks contributes to a comprehensive assessment of the effectiveness of model editing methods.

The construction of \benchmarkname~is divided into two steps: (1) Collect real fact triples; (2) Create diverse tasks using the collected triple.
To ensure the data quality of \benchmarkname~and its reflection of the real world, we conducted multiple rounds of manual verification and correction in various aspects. 
For more details, please refer to Appendix \ref{app:benchmark_details}.

\subsection{Collect Fact Triples }
\label{subsec:choose fact triples}

To obtain real-world fact triples, we collect data from Wikidata \citep{vrandevcic2014Wikidata} and DBpedia \citep{auer2007DBpedia}, both of which are continuously updated databases. We aim to enhance the diversity of \benchmarkname~by collecting knowledge from a variety of knowledge bases.

Specifically, we initially identified equivalent relations in Wikidata and DBpedia. Subsequently, non-informative relationships such as IDs were discarded. Then, we collect triplets associated with these relations from Wikidata and DBpedia.

After obtaining the triplets, we further filter them to avoid potential ambiguity issues, see Appendix  ~\ref{sec:Datafilter} for details.

Finally, to ensure the quality of the triplets we obtained, we randomly selected 100 triplets and manually examined their correctness. The results indicate that 96\% of the triplets are correct, which shows that our process for obtaining and filtering triplets is acceptable.

\subsection{Generate Data Based on Templates} 
\label{subsec:Generate Data Based on Templates}
We create templates for each type of task to transform fact triples into queries for various tasks. We employ ChatGPT in the generation process to mitigate expensive labor costs following previous works \citep{petroni2019language, yin2023alcuna}. After generating the results, we conduct manual checks to ensure the accuracy and alignment with our intentions.

For single-hop questions, we prompt ChatGPT to generate question templates based on the description of each relationship that used in fact triples (e.g., head of government), incorporating placeholders. Then we replace these placeholders with subjects to generate questions from the templates.

For multi-hop questions, we employ ChatGPT to concatenate multiple consecutive triplets into a single question. Inspired by \citet{petroni2019language}, to distinguish between the differences in model decomposition ability and knowledge it knows, we decompose queries to obtain multi-turn dialogue.

To ensure the accuracy of templates, we incorporate manual verification to ensure that the templates align with the meaning of the relationships. We found that 97.4\% of templates are accurate, we then manually performed multiple rounds of correction and rechecking, ensuring that the correctness rate of the templates reached 100\%.

Finally, following previous work \citep{yin2023alcuna}, we employ manual sampling and verification techniques to ensure the accuracy of \benchmarkname. We combine the templates and relation triplets and manually check the credibility of the generated sentences. The results show that 97.5\% of the sentences were credible, demonstrating the reliability of the entire process.

\section{Benchmark Analysis}
\subsection{Comparison}
See Table \ref{dataset_cmp} for a comparison between \benchmarkname~ and previous benchmarks. \benchmarkname~includes all categories seen in previous benchmarks, and we propose additional data categories. Moreover, the number of entries far exceeds those in most of the previous benchmarks. Finally, \benchmarkname~ originates from two distinct knowledge bases, making it more comprehensive compared to previous datasets.

Compared to the previously multi-hop dataset containing factual knowledge, MQuAKE-T \citep{zhong2023MQuAKE}, \benchmarkname~is larger and includes more relationships, making it more comprehensive. See Appendix \ref{app:detail_cmp_datasets} for details on the comparison and further comparisons.

\subsection{Statistics}
   \benchmarkname~ consists of two parts: single-hop data and multi-hop data, both sourced from Wikidata and DBpedia. Table \ref{tab:task_types} presents the statistical data for different tasks in the dataset.  
   Please refer to Appendix \ref{app:example_sta_for_dataset} for examples and more detailed statistics.
 \begin{table}[!ht]
    \centering
    \begin{tabular}{ccc}
    \hline
    \textbf{Task} & \textbf{Type} & \textbf{Total} \\ \hline
    
    \multirow{6}{*}{Single-hop} & Single-hop QA & 20,000 \\ 
                                & Sentence Completion & 20,000 \\ 
                                & Cloze & 20,000 \\ 
                                & Multiple-Choice & 20,000 \\ 
                                & Fact Check & 20,000 \\ 
                                & Locality & 20,000 \\ \hline
    \multirow{8}{*}{Multi-hop}  & Multi-hop QA in 2 hops & 1,000 \\ 
                                & Dialogue in 2 hops & 1,000 \\ 
                                & Multi-hop QA in 3 hops & 1,000 \\ 
                                & Dialogue in 3 hops & 1,000 \\ 
                                & Multi-hop QA in 4 hops & 1,000 \\ 
                                & Dialogue in 4 hops & 1,000 \\ 
                                & Multi-hop QA in 5 hops & 1,000 \\ 
                                & Dialogue in 5 hops & 1,000 \\ \hline
    
    \end{tabular}
    \caption{Task types and statistics}
    \label{tab:task_types}
\end{table}

\section{\methodname: A Model Editing Method For Real-World Applications}
\begin{figure*}[!ht]
    \centering
    \includegraphics[width=\linewidth,height=0.55\linewidth]{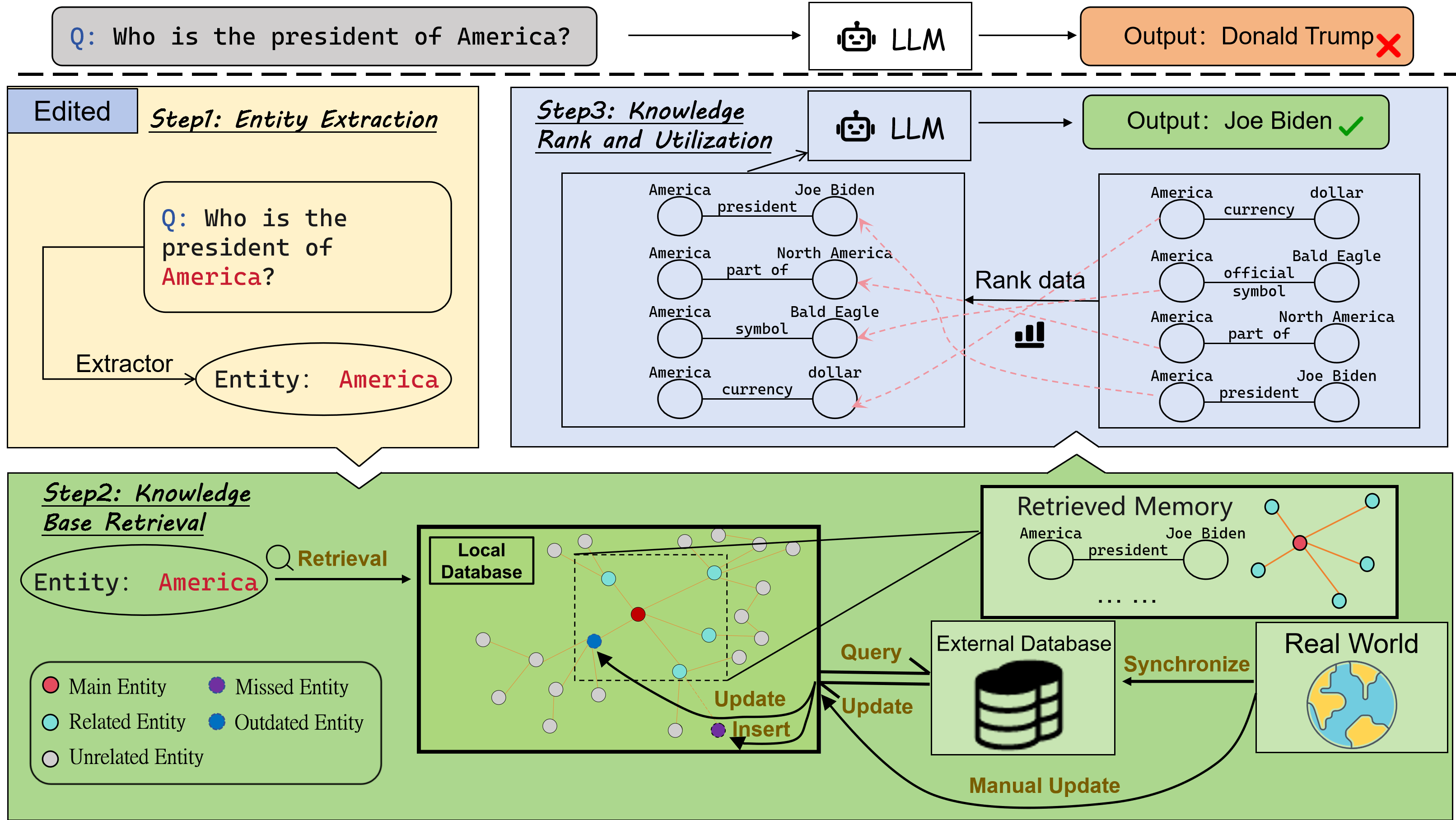}
    \caption{An overview of \methodname. \methodname~initially extracts key entities from the question. Subsequently, it retrieves the knowledge base for facts related to entities. Then ranks applicable knowledge items and utilizes in-context learning to modify the model's output. Additionally, we update knowledge from external databases and the real world to ensure that the local knowledge base reflects real-world changes.}
   \label{pic:model}
\end{figure*}

To accommodate practical model editing and meet the need for utilizing real-world facts and adapting to diverse tasks, We propose \methodname~(\textbf{S}tructured \textbf{K}nowledge retrieved by \textbf{E}xact \textbf{M}atching and reranking \textbf{E}diting), a model editing method that first incorporates a caching system to efficiently store and retrieve real-world knowledge, which is collected from diverse data sources. Besides, \methodname~utilizes entity extraction to exclude irrelevant content brought by diverse input formats, enhancing its performance across various tasks.

\subsection{Overview}
The overview of \methodname~is shown in Figure \ref{pic:model}.
\methodname~consists of three main components: 

To avoid the influence of varied input formats, \textbf{Entity Extraction} extracts key entities from the input for subsequent retrieval. 
\textbf{Knowledge Base Retrieval} queries external knowledge bases for world knowledge related to the entities extracted by Entity Extraction and caches the results in the local structured knowledge base.
\textbf{Knowledge Rank and Utilization} utilizes the results retrieved by Knowledge Base Retrieval to correct the model's output. Please refer to Appendix \ref{sec:model_datail} for details.

    \subsection{Entity Extraction}
    \label{sec:entity_extraction}

To handle complex real-world tasks, \textbf{entity extraction} aims to extract key entities \( e \) from the input \( I \), while ignoring irrelevant content (e.g., extracting "America" from "Who is the president of America?"), thus excluding the influence of input forms and preparing the retrieval entity for the next step.
Specifically, to ensure system robustness, \methodname~prompt uses LLMs to extract key entities \( e \) from the input \( I \).

     \subsection{Knowledge Base Retrieval}
    \label{sec:know_base_retrival}

 To accurately retrieve facts related to the entity extracted by entity extraction \( e \), we use a knowledge graph (KG) \( KG = \{ (s, r, o) \mid s, o \in E, r \in R \} \), where \( E \) is the set of entities and \( R \) is the set of relations. The retrieval step involves extracting a subgraph \( G' \) from the knowledge base \( G \), where \( G' = \{(s', r', o') \mid (s', r', o') \in G \text{ and } s' = e\} \).

Specifically, in knowledge base retrieval, similar to the principle of locality in \textbf{computer caching systems}, some knowledge may reappear across multiple queries \citep{jin2024ragcache}. Therefore, introducing a caching system can reduce the number of data queries and improve system efficiency.
Inspired by this, we utilize a fast-slow table mechanism in the knowledge base. The slow table is \( G \)  and the fast table is \( G''  = \{(s'', r'', o'') \mid (s'', r'', o'') \in G \text{ and } s'' \in E \} \), respectively. It's important to note that \( G'' = \varnothing \) initially.

Initially, we retrieve facts related to the previously extracted entity \( e \) in the \( G'' \). If related facts are not situated in \( G'' \), we then search in the slow table \( G \) and update \( G'' \) with the retrieval results. Inspired by the principle of locality in operating systems, we not only update retrieval results in \( G'' \) but also update the triples related to the retrieval results in \( G'' \).

To ensure consistency between \( G \) and \( G'' \), we build the \( G'' \) through two primary methods. First, if the retrieval in the local database yields \( G' = \varnothing \), indicating that the entity is absent from the \( G'' \), we retrieve the extracted entity from \( G \) in the same approach as \( G'' \), then load the retrieval results in the \( G \) to update our \( G'' \). 
Since \( G \) such as Wikidata continues to update to remain current with the real world, the \( G'' \) sourced from \( G \) essentially serves as a cache of real-world facts, allowing the \( G'' \) to reflect real-world facts. 
Second, benefiting from the use of structured knowledge bases, \( G'' \) can also be manually updated to directly incorporate knowledge from the real world as an additional source of information.

To keep the \( G'' \) current with the \( G \), we will update the \( G'' \) regularly. In the process of updating, for any update \((s,r,o_{new})\), if outdated knowledge \((s,r,o_{old}) \) exists in \( G'' \) and \(o_{new} \neq o_{old}\), we replace \(o_{old}\) with \(o_{new}\); otherwise, we directly insert the new knowledge. This strategy allows us to adapt to the evolution of facts, such as the transition of the U.S. President from Obama $\rightarrow$ Trump $\rightarrow$ Biden.

     \subsection{Knowledge Rank and Utilization}

After retrieving the graph \( G' \), we use the pre-trained Contriever \citep{izacard2021unsupervised} model to calculate embeddings for the triples in \( G' \).  Subsequently, we retain the facts that are closest to the query in the embedding space, which means the facts most relevant to the query.

 Inspired by \citet{zheng2023can}, we use in-context learning to modify the model's output. Specifically, we integrate the triples into the input and prompt the model to utilize the triples in responding to queries. Such in-context learning enables us to tackle various tasks economically and ensures its effectiveness in models of varying sizes.

\section{Experiment}
\label{sec:main_exp}
\subsection{Metrics}

We use the following metrics to evaluate whether editing achieves our goal in Section \ref{sec:problemsDefinition}.

\paragraph{Accuracy}
    To calculate accuracy, we instruct the model to generate responses for tasks and evaluate whether they match the gold answers exactly. The resulting average accuracy is then recorded as \textbf{exact match (EM)}.
    
\paragraph{Locality}

Locality measures whether an editing method will influence 
 irrelevant knowledge. We utilize \textbf{drawdown (DD)} \citep{mitchell2021fast,mitchell2022memory}~to compute performance degradation and employ \textbf{Neighborhood KL divergence (NKL)} \citep{hoelscher2023detecting}~to measure whether the model is significantly affected.

\paragraph{\scorename}

\label{sec:scorename}
To meet the demands of practical application, model editing needs to consider both accuracy and side effects while also adapting to scenarios. To assess this capability, we propose the metric \textbf{\scorename}~(\textbf{S}tatistical and \textbf{U}nbiased \textbf{R}eal-world \textbf{E}valuation) to estimate the performance of edited models in real-world scenarios. We define \scorename~as follows:
\begin{equation}
\setlength{\abovedisplayskip}{2pt} 
\setlength{\belowdisplayskip}{2pt}
SURE=aEM^{\alpha } -  bDD^{\beta }
\nonumber 
\end{equation}
The parameters $a$ and $b$ denote the ratio of the data used to evaluate the two metrics, $\alpha$ and $\beta$ are used to characterize the importance of EM and DD, which are adjusted according to specific tasks.
 See Appendix \ref{app:detail_analysis_of_score} for a more detailed analysis of its motivation and advantage.

\paragraph{Efficiency}
Efficiency measures the time and GPU space consumed by the model editing methods. Following \citet{yao2023editing}, we measure efficiency in both \textbf{time consumption (Ti)} and \textbf{memory requirements (Me)}.

\subsection{Baselines}
Following \citet{yao2023editing}, we compare \methodname~with parameter-modifying methods, including FT and MEMIT \citep{meng2022mass}, as well as parameter-preserving methods, including MeLLo \citep{zhong2023MQuAKE}, and IKE \citep{zheng2023can}. FT is the most classic and straightforward model-editing method. MEMIT is currently considered a state-of-the-art method among parameter modification methods. IKE and MeLLo, much like \methodname, leverage a knowledge base and in-context learning.
Implementation details can be found in Appendix \ref{sec:Baselines}.

\begin{table*}[!ht]
    \renewcommand{\arraystretch}{0.9}
    \centering
    \tabcolsep=0.09cm
    \begin{tabular}{ccccccccccccc}  

      \hline
        \multirow{2}*{Model} & \multirow{2}*{Method}& \multirow{2}*{\textbf{\scorename$\uparrow$}} &\multicolumn{6}{c}{Accuracy}  &  \multicolumn{2}{c}{Locality}  &  \multicolumn{2}{c}{Efficiency}  \\
		\cline{4-9}
		\cline{10-11}
		\cline{12-13}
        ~&  ~& ~&  \textbf{EM.$\uparrow$ }&  QA.$\uparrow$&
        Com.$\uparrow$ & Clo.$\uparrow$ & Cho.$\uparrow$ & FC.$\uparrow$   & \textbf{DD. $\downarrow$ }&\textbf{ NKL. $\downarrow$} &\textbf{ Ti. $\downarrow$ }&\textbf{ Me.  $\downarrow$ }\\
     
        \hline\multirow{6}*{GPT2-XL} & Base & - & $19.83$ & $8.00$ & $7.11$ & $3.63$ & $34.25$ & $46.16$ & - & - & $0.18$ & $9.12$ \\
         ~ & FT & $12.75$ & $22.72$ & $11.82$ & $10.26$ & $9.96$ & $33.58$ & $47.96$ & $9.97$ & $1.33$ & $2.12$ & $12.43$ \\
        ~ & MEMIT& $20.87$ & $20.87$ & $7.31$  & $7.14$ & $6.67$ & $34.22$ & $49.04$ & $\textbf{0.00}$ & $1.29$  & $13.6$ & $11.85$\\
        ~ & MeLLo & $-53.67$ & $30.90$ & $71.42$ & $0.24$ & $0.09$ & $33.72$ &  $49.01$ & $84.57$ & $1.32$ & $1.43$ & $17.43$\\
        ~ & IKE & $37.32$ & $50.51$ & $62.05$ &$54.82$ & $48.96$ & $36.09$ & $50.64$ & $13.19$ & $1.25$ & $0.75$ & $14.26$ \\
        ~ & \methodname& $\textbf{65.80}$ & $\textbf{65.80}$ & $\textbf{85.12}$  & $\textbf{70.60}$ & $\textbf{78.45}$ & $\textbf{38.33}$ & $\textbf{56.51}$ & $\textbf{0.00}$ & $\textbf{1.09}$ & $\textbf{0.23}$ & $\textbf{11.52}$ \\
       \hline
      
        \multirow{6}*{GPT-J} & Base & - & $23.36$ & $11.86$ & $12.02$ & $11.52$ & $35.34$ & $46.08$ & - & - & $0.35$ & $26.57$ \\
        ~ & FT & $25.21$ & $26.59$ & $13.69$ & $13.38$ & $13.39$ & $40.74$ & $51.72$ & $1.38$ & $1.76$ & $3.27$ & $34.81$ \\
        ~ & MEMIT  & $45.85$ & $45.85$ &
         $49.51$  & $41.14$ & $43.51$ & $\textbf{46.62}$ & $48.49$  & $\textbf{0.00}$ & $1.86$ & $13.8$ & $29.84$\\
        ~ & MeLLo & $28.42$ & $55.74$ & $72.20$ & $48.35$ & $72.95$ & $21.81$ & $\textbf{63.41}$ & $27.33$ & $1.54$ & $2.42$ & $33.38$ \\
        ~ & IKE & $58.62$ & $70.04$ & $87.00$ & $\textbf{82.35}$ & $82.27$ & $46.32$ & $52.26$ & $11.42$ & $\textbf{1.28}$ & $0.97$ & $31.53$ \\
        ~ & \methodname & $\textbf{73.93}$  & $\textbf{73.93}$ & $\textbf{97.03}$ & $79.63$ & $\textbf{87.02}$ & $46.01$ & $59.97$ & $\textbf{0.00}$ & $1.39$  & $\textbf{0.49}$ & $\textbf{28.17}$ \\
        \hline
        \multirow{6}*{Llama2} & Base & - & $32.20$ & $15.82$ & $15.78$ & $16.02$ & $48.91$ & $64.45$ & - & - & $0.33$ & $30.04$ \\
        ~ & FT & $34.31$ & $41.80$ & $30.05$ & $29.08$ & $29.22$ & $60.57$ & $60.07$ & $7.49$ & $2.55$ & $5.18$ & $38.92$ \\
        ~ & MEMIT & $48.03$ & $48.39$ & $41.16$ & $40.26$ & $41.47$ & $61.00$ & $58.08$ & $0.36$ & $2.83$ & $13.2$ & $33.52$ \\
        ~ & MeLLo & $36.38$ & $66.26$ & $68.56$ & $36.95$ & $69.26$ & $78.17$ & $78.35$ & $29.88$ & $2.72$ & $2.45$ & $38.66$ \\
        ~ & IKE & $71.38$ &  $\textbf{91.42}$ & $97.72$ & $\textbf{90.11}$ & $\textbf{95.76}$ & $\textbf{95.10}$ & $78.42$  & $20.04$ &  $2.48$ & $1.08$ &$35.17$ \\
        ~ & \methodname & $\textbf{90.54}$  & $90.54$ & $\textbf{98.61}$ & $83.04$ & $90.27$ & $93.73$ &$\textbf{87.07}$ & $\textbf{0.00}$ & $\textbf{2.12}$ & $\textbf{0.45}$  & $\textbf{31.83}$ \\
      \hline   
      
      \multirow{4}*{\makecell[c] {GPT-\\3.5-turbo}} & Base  & - & $40.11$ & $18.76$ & $19.65$ & $17.17$ & $73.73$ & $71.22$ & - & $*$ & $0.81$ & $*$ \\

        ~ & MeLLo & $56.58$ & $73.75$ & $70.51$ & $57.16$ & $76.37$               
                    & $82.78$ & $81.92$ & $17.16$ & $*$ & $2.92$ & $*$ \\
        ~ & IKE & $76.45$ & $89.53$  & $92.81$ & $\textbf{89.72}$ & $\textbf{90.88}$ & $90.41$ & $83.85$  & $13.08$ & $*$  & $1.47$  & $*$\\
        ~ & \methodname &  $\textbf{91.76}$  &  $\textbf{91.76}$ & $\textbf{98.07}$ & $84.78$ & $89.45$ & $\textbf{99.04}$ & $\textbf{87.40}$  & $\textbf{0.00}$ & $*$  & $\textbf{1.03}$  & $*$ \\
      
      \hline
    \end{tabular} 
    \caption{
    \label{mainresult}Main result on \benchmarkname.     
  "Com.", "Clo.", "Cho." and "FC." stands for completion, cloze, choose, and fact-checking, respectively.
  Ti (s) includes both editing and generating time in Wall clock time and Me (GB) is calculated by measuring the maximum required GPU VRAM. To maintain brevity, the multiplier of $\times 10^{-4}$ has been excluded for the NKL metric. Since DD and NKL are calculated relative to the unedited model, the unedited model does not have these metrics. 
  "-": Since DD and NKL are computed relative to the unedited model, and SURE's calculation depends on DD, these three metrics are meaningless for the unedited model.
  "$*$": The computation of NKL and ME metrics for GPT-3.5-turbo is impractical due to its utilization via API calls.
}
\end{table*} 

\subsection{Main Result}

    Table \ref{mainresult} shows results on \benchmarkname. We experiment with all methods on GPT2-XL \citep{solaiman2019release}, GPT-J (6B) \citep{wang2021gpt}, Llama2 \citep{touvron2023Llama2}, and utilize in-context learning based methods on GPT-3.5-turbo \citep{ouyang2022training}.

        Firstly, we scrutinize the results on Llama2, which is the largest model we can employ all model editing methods. \textbf{FT} and \textbf{MEMIT}, did not perform well in our experiments, which may be due to the editing process not specifically targeting the model's generative capability. \textbf{MeLLo} has a higher EM score than FT and MEMIT, but its DD is also the highest, indicating its pronounced side effects, which leads to a low \scorename. Both \textbf{IKE} and \textbf{\methodname} obtained an EM above 0.9. However, IKE also has presented side effects that consequently decreased its \scorename. \methodname~uniquely maintains a high EM and simultaneously ensures a low DD, thus demonstrating superior practicality compared to other methods.

        To test the impact of model size, we experiment with various model sizes. \methodname~excels across all, while some other methods fail on small models.        
        These model editing methods also require diverse amounts of time and GPU space. MeLLo, due to its long in-context learning process, consumes the most time in the RAG methods. 
        MEMIT demonstrates strong capability and low side effects, but it is more time-consuming. Additionally, previous work \citep{yao2023editing} has pointed out that while MEMIT can perform batch editing, its effectiveness tends to decrease as the batch size increases. In contrast, models based on RAG and in-context learning (MeLLo, IKE, and SKEME) can easily handle batch editing without decreasing performance. 
        Overall, \methodname~proves effectiveness across model sizes while consuming less additional time and GPU space.

            It appears that all methods performed poorly on certain tasks. This further validates the meaningfulness of constructing data in various forms.
            On the completion task, although the base model performed similarly to QA and Cloze, the edited model's accuracy was significantly lower than QA and Cloze. It indicates that the method's generalization performance still needs to improve.

\section{Analysis}
\label{sec:Analysis}

Considering the complexity of real world, besides single fact edits, we design a series of research questions (RQs) to evaluate the method's ability to edit multiple facts. We discuss \textbf{fact transitions (RQ1)}, \textit{e.g., the U.S. President transitioning from Obama $\rightarrow$ Trump $\rightarrow$ Biden}, \textbf{fact inference (RQ2)}, \textit{e.g., inferring the fact “the First Lady of the U.S. is Jill Biden” from the given facts “the U.S. President is Biden” and “Biden's spouse is Jill Biden”}, \textbf{fact with substantial quantity (RQ3)}, \textit{e.g., needing to update thousands of facts} and \textbf{fact from various benchmarks (RQ4)}, \textit{e.g., facts from other datasets.}

\setlength{\textfloatsep}{7pt}
\subsection{RQ1: How Do Methods Handle Transitions Between Facts?}

Many facts in the real-world transition require multiple edits to the same fact. To evaluate the effectiveness of repeated edits, we perform multiple updates for each fact and tested the accuracy of the edited model.
\begin{figure}[!ht]
    \centering
    
    \begin{tikzpicture}[scale=0.65]
        \begin{axis}[
            ymin=0, ymax=1,  
            xlabel={Edited Times},
            ylabel={EM},
            legend style={at={(0.5, -0.25)}, anchor=north, legend columns=5},
            enlarge y limits={upper,value=0.15}, 
            xtick={1, 2, 5, 10}, 
            xticklabels={1, 2, 5, 10},  
            x tick label style={/pgf/number format/fixed}, 
            xtick distance=1, 
        ]

        \addplot [mark=o] coordinates {
            (1, 0.307) (2, 0.245) (5, 0.190) (10, 0.164)
        };
        \addplot coordinates {
            (1, 0.405) (2, 0.231) (5, 0.133) (10, 0.103)
        };
        \addplot coordinates {
            (1, 0.684) (2, 0.378) (5, 0.214) (10, 0.170)
        };
        \addplot [color=orange, mark=x]  coordinates {
            (1, 0.974) (2, 0.529) (5, 0.378) (10, 0.301)
        };
        \addplot coordinates {
            (1, 0.986) (2, 0.986) (5, 0.986) (10, 0.986)
        };

        \legend{FT,  MEMIT, MeLLo, IKE, \methodname~ (Ours)}
        \end{axis}
    \end{tikzpicture}
    \caption{Result of RQ1. The x-axis indicates the number of edits to the same fact.}
    \label{fig:RQ1}
\end{figure}
Figure \ref{fig:RQ1} shows the experimental results. The results indicate that even with only two edits to the same fact, the accuracy of all other methods substantially declines. 
Parameter-modifying methods can lead to divergence from the initial model parameters and subsequent performance decline due to repeated adjustments of specific parameters.
For methods not to modify the parameters, failure to update the knowledge base may result in conflicts between existing and newly added facts. 

\methodname~ uses a structured knowledge base to facilitate precise updates, making iterative updates possible. \methodname~ is the only one capable of handling iterative updates.

\subsection{RQ2: Can The Edited Model Infer New Facts Based on Given Information?}
 The edited model should be able to make further reasoning based on the edited facts. Following previous research \citep{zhong2023MQuAKE}, We employ multi-hop questions to evaluate this capability of the model.

\begin{table}[!ht]
    \renewcommand{\arraystretch}{0.9}
    \centering
    \begin{tabular}{ccccc}
      \hline
        \multirow{2}*{\textbf{Method}}& \multicolumn{4}{c}{\textbf{Multi-hop QA} } \\
    \cline{2-5}
        ~ &2-hops&3-hops&4-hops&5-hops\\
      \hline
         Base   & 0.145 & 0.135 & 0.112 & 0.079 \\
         FT  & 0.223 & 0.362 & 0.231 & 0.128 \\
         MEMIT   & 0.176 & 0.247 & 0.136 & 0.060 \\
         MeLLo  & 0.270& 0.227& 0.167 & 0.073 \\
         IKE   & 0.332 & 0.237 & 0.220 & 0.159 \\

      \methodname & \textbf{0.960}&\textbf{0.786} &\textbf{0.427} & \textbf{0.167} \\

      \hline
        \multirow{2}*{\textbf{Method}}& \multicolumn{4}{c}{\textbf{Dialogue} } \\
      
    \cline{2-5}
        ~ &2-hops&3-hops&4-hops&5-hops\\
      \hline
         Base & 0.119 & 0.118 & 0.116 & 0.082  \\
         FT & 0.190 & 0.216 & 0.152 & 0.133  \\
         MEMIT & 0.238 & 0.220 & 0.148 & 0.126  \\
         MeLLo & 0.353 & 0.295 & 0.193 & 0.111  \\
         IKE & 0.229 & 0.235 & 0.207 & \textbf{0.188}   \\
         \methodname & \textbf{0.946} & \textbf{0.757} & \textbf{0.390} & 0.181 \\
        
      \hline
    \end{tabular}
    \caption{
    \label{tab:rq4} Result of RQ2. \methodname~manifests significant improvements compared to previous approaches, however, it still fails to address the issue when \( hops \geq 4 \).  }
\end{table}

Table~\ref{tab:rq4} presents the results for this task. We can observe that all methods, except for \methodname, performed poorly. Traditional retrieval-based models struggle to find answers to multi-hop questions, and other methods do not enable the model to infer based on edited facts.

\subsection{RQ3: Can Methods Handle Updates with a Substantial Amount of Facts?}

In the real world, there are numerous updates to facts and a practical model editing method should be able to update a vast quantity of knowledge in the model.

\begin{figure}[!h]
    \centering
  \begin{tikzpicture} [scale=0.65]
\begin{axis}[
            xlabel={Number of facts},
            ylabel={EM},
            xmode=log, 
            legend style={at={(0.5, -0.25)}, anchor=north, legend columns=5},
            enlarge y limits={upper,value=0.1}
        ]
\addplot [mark=o] coordinates{
(1, 0.301) (10, 0.308) (100, 0.172) (1000, 0.046) (10000, 0.003) (100000, 0.000)};
\addplot coordinates{
(1, 0.420) (10, 0.420) (100, 0.390) (1000, 0.000) (10000, 0.000) (100000, 0.000)};
\addplot coordinates{
(1, 0.686) (10, 0.624) (100, 0.619) (1000, 0.620) (10000, 0.618) (100000, 0.616)};
\addplot [color=orange, mark=x] coordinates{
(1, 0.977) (10, 0.974) (100, 0.973) (1000, 0.971) (10000, 0.970) (100000, 0.970)};
\addplot coordinates{
(1, 0.986) (10, 0.986) (100, 0.986) (1000, 0.986) (10000, 0.986) (100000, 0.986)
};

\legend{FT,MEMIT,MeLLo,IKE,\methodname~ (Ours)}
\end{axis}
\end{tikzpicture}
    \caption{Result of RQ3. The x-axis represents the number of edited facts.}
    \label{fig:RQ2}
\end{figure}

Figure \ref{fig:RQ2} shows the experimental results. 
As the quantity of modified facts increases, parameter-modifying methods progressively shift away from their original state, resulting in a notable performance decline. In contrast, other methods, exhibit only slight declines in performance and can handle updates to a large number of facts.

\subsection{RQ4: Can Methods Generalize Across Facts in Various Benchmarks?}

To demonstrate the general applicability of the editing method, we select several datasets that are widely used to evaluate LLMs' understanding of the world, and subsequently evaluate model editing methods on them. Please refer to Appendix \ref{sec:other_benchmarks} for details about these datasets.

\begin{table}[!ht]
    \renewcommand{\arraystretch}{0.9}
    \centering
    \begin{tabular}{ccccc}
      \hline
        \textbf{Method}& \textbf{TQA}& \textbf{NQ}& \textbf{FEVER}  & \textbf{Vi}   \\

      \hline
         Base & $0.698$ & $0.191$ & $0.792$ & $0.397$\\
         FT & $0.362$ & $0.274$ & $0.646$ & $0.228$\\
         MEMIT & $0.449$ & $0.632$ & $0.724$ & $0.461$ \\
         MeLLo  & $0.811$ & $0.633$ & $0.872$ & $0.720$ \\
         IKE & $0.962$ & $\textbf{0.980}$ & $0.954$ & $\textbf{0.964}$\\
      \methodname & $\textbf{0.984}$ & $0.964$ & $\textbf{0.987}$ & $0.956$ \\
      \hline
    \end{tabular}
    \caption{
    \label{tab:rq3}Result of RQ4. 
    "TQA", "NQ" and "Vi" stands for triviaQA, Natural Questions, and VitaminC, respectively.    
    All accuracies are calculated based on exact match rates.
    }
\end{table}

Table \ref{tab:rq3} shows that \methodname~ consistently improves the model's performance irrespective of the benchmarks, demonstrating the robust versatility and scalability of \methodname.

\section{Conclusion}

We introduce the practicality requirement for model editing and created a novel benchmark \benchmarkname, which embodies practicality with factual data and diverse tasks. We propose a model editing method, \methodname, that proves effective across various LLMs and tasks. The experiments demonstrate that previous model editing methods struggle in dealing with real-world challenges, while \methodname~successfully addresses these challenges. We hope that our work will advance the field of model editing and inspire further research in this area.

\section*{Acknowledgments}
Thanks for the insightful comments from reviewers. This work is supported by the Natural Science Foundation of China (No. U21B2009, 62406015) and the Beijing Institute of Technology Science and Technology Innovation Plan (No. 23CX13027).
\section*{Limitations}

The data in \benchmarkname~is limited to a monolingual scope and does not include multilingual data. We posit that the inclusion of multilingual data can further align with the real world, and we leave this as a potential area for future work.

\section*{Ethics Statement}

We ensure that the collection of \benchmarkname~is done in a manner consistent with the terms of use stipulated by its sources and the intellectual property rights of the original authors. We make sure that individuals involved in the collection process are treated fairly, including ensuring their voluntary participation and informed consent. Due to the dynamic nature of the real world, certain knowledge contained in \benchmarkname~may become outdated, rendering it no longer reflective of the latest world conditions.

\bibliography{anthology,custom}
\bibliographystyle{acl_natbib}

\appendix

\section{Implementation Details}
\subsection{\benchmarkname~ Construction Details }
This section provides a detailed description of how we constructed \benchmarkname.
\label{app:benchmark_details}
\subsubsection{Collect Fact Triples}

To enhance the diversity of \benchmarkname~ and reduce bias from a single data source, We use WikiData SPARQL query\footnote{\url{https://query.wikidata.org}} and  DBpedia SPARQL query\footnote{\url{https://dbpedia.org/sparql}} to collect data from Wikidata and DBpedia.

Firstly, using the code in Figure \ref{fig:data_relationship}, we query for relationships with equivalent meanings in Wikidata and DBpedia, such as "birth place" (from DBpedia) and "place of birth" (from Wikidata). They are connected through the relationship \textit{equivalentProperty}. Subsequently, we filter out relationships like identifiers, where their objects are typically composed of irregular and meaningless combinations of letters and numbers.

Next, based on the obtained $r$, we use code in Figure \ref{fig:query_sro_wiki} and \ref{fig:query_sro_db} to collect triples $(s, r, o)$ from Wikidata and DBpedia. We filter out triples that may cause ambiguity, including two aspects: two different items having the same name or a specific entity's relation corresponding to multiple objects. Relevant discussions can be found in Appendix \ref{sec:Datafilter}.
\begin{figure*}[!t]
\centering
\begin{minipage}{\textwidth}
\begin{lstlisting}[]
SELECT ?DBpediaProp ?itemLabel ?WikidataProp
WHERE
  {
    ?DBpediaProp  owl:equivalentProperty  ?WikidataProp .
          FILTER ( CONTAINS ( str(?WikidataProp) , 'wikidata' ) ) .
    ?DBpediaProp    rdfs:label    ?itemLabel .
          FILTER (lang(?itemLabel) = 'en')
  }
ORDER BY  ?WikidataProp
\end{lstlisting}
    \caption{Sparql code used to query equivalent relations.}
    \label{fig:data_relationship}
\end{minipage}
\end{figure*}

Lastly, we manually extract 100 distinct triples and verify them against other data sources such as government websites to ensure the accuracy and real-world relevance of our collected triples. The results show that 96\% of the data is correct. Hence, we can infer that \benchmarkname~can reasonably reflect real-world scenarios.

\begin{figure*}[!t]
\centering
\begin{minipage}{\textwidth}
\begin{lstlisting}[]
SELECT DISTINCT ?subject ?object ?subjectLabel ?objectLabel (COUNT(distinct ?r) AS ?relationCount)
WITH {
SELECT DISTINCT ?subject ?object ?subjectLabel ?objectLabel
    WHERE {
      ?subject wdt:{item} ?object.
      # ?subject wdt:P31 wd:Q5.
  ?subject rdfs:label ?subjectLabel.   
     FILTER(LANG(?subjectLabel) = "en").
  OPTIONAL {?object rdfs:label ?objectLabel.}  
     FILTER(LANG(?objectLabel) = "en").
    }
LIMIT {limit}
OFFSET {offset}
} AS %SUB

WHERE{
  INCLUDE %SUB
  ?subject ?r []
}
GROUP BY ?subject ?object ?subjectLabel ?objectLabel
\end{lstlisting}
    \caption{Code used to query triples from Wikidata.}
    \label{fig:query_sro_wiki}
\end{minipage}
\end{figure*}

\begin{figure*}[!t]
\centering
\begin{minipage}{\textwidth}
\begin{lstlisting}[]
PREFIX rdfs: <http://www.w3.org/2000/01/rdf-schema#>
SELECT ?subject ?subjectLabel ?object ?objectLabel 
WHERE {
  ?subject <{property_url}> ?object.
  # ?subject <http://dbpedia.org/ontology/primeMinister> ?object.
  ?subject rdfs:label ?subjectLabel.
     FILTER(LANG(?subjectLabel) = "en").
 OPTIONAL { ?object rdfs:label  ?objectLabel. FILTER(LANG(?objectLabel) = "en"). }
}
\end{lstlisting}
    \caption{Code used to query triples from DBpedia.}
    \label{fig:query_sro_db}
\end{minipage}
\end{figure*}

\subsubsection{Data Filter}
\label{sec:Datafilter}

Ambiguity issues involve two aspects: different entities sharing the same name and a specific entity's relation corresponding to multiple objects.

For the former scenario, one example is: \emph{Hope Springs} could refer to a movie from 2012 (Q327214 in Wikidata)\footnote{\url{https://www.Wikidata.org/wiki/Q327214}}, but can be a movie from 2003 as well (Q596646 in Wikidata)\footnote{\url{https://www.Wikidata.org/wiki/Q596646}}. So when asking \emph{Who is the director of Hope Springs?}, there are multiple correct options.

An example of the latter scenario is: a person may have multiple children, so there are multiple correct answers when asking for their children's names.

We believe that the above two scenarios are simpler compared to questions with only one answer. Therefore, for easier implementation and to focus on more fundamental phenomena, we excluded data in the dataset containing instances of the above situations.

\subsubsection{Generate Data Based on Templates}
Following previous work, after obtaining triples, we need to construct relationship templates to build our entire dataset. For single-hop data, we use the following triple as an example to illustrate our entire construction process: \textit{(subject, relation, object)} = \textit{(America, head of government, Biden)}.

We construct several templates for each relation for each task. For instance, when \( (r=\text{head of government}) \), the template for the QA task might be ``Who is the head of government in \{\}?'', and the template for the completion task might be ``In \{\}, the head of government is''. During usage, we replace \{\} with the subject to obtain the QA and completion task queries: ``Who is the head of government in America?'' and ``In America, the head of government is''. We expect the model to answer with the object Biden.

Following previous work, we use ChatGPT to assist us in constructing templates. We provide each relation name and a brief description and use few-shot learning to ensure that the templates it constructs meet our requirements. We require it to construct 3-4 different templates for each relation to test different template generation methods for their generalization performance.

For multi-hop questions, consider a chain \( C = [(s_1, r_1, o_1), (o_1, r_2, o_2), \ldots] \), where the object of each triple is the subject of the next triple, i.e., \( o_i = s_{i+1} \). When \( o_1 \) changes, the entire chain undergoes a ripple effect of changes, and we expect the model to answer the updated last object.

The multi-hop task contains two subtasks: multi-hop questions and dialogues. Taking a triple chain as an example: \textit{[(America, head of government, Biden), (Biden, spouse, Jill Biden)]}, we can write multi-hop questions templates like: ``Who is the spouse of the head of government in \{\}?'', and then generate the question ``Who is the spouse of the head of government in America?''. We prompt ChatGPT to construct corresponding templates.   
As for the dialogue task, we use the QA questions from single-hop questions and replace the intermediate objects with appropriate pronouns. For example, a dialogue could be: \textit{Q: Who is the head of government in America? A: < LLM answer >. Q: Who is his spouse? A: < LLM answer >.} The intuition is that multi-hop questions are more common but require the model to have good reasoning abilities, while dialogues focus on testing if the model can know updated facts.

Afterward, we manually check if the templates are correctly constructed. We primarily focus on whether the templates' meanings are appropriate. A common mistake is reversing the relationship between subject and object, for example, \textit{(subject, owner of, object)} means ``subject is the owner of object'', but ChatGPT might reverse this relationship. We manually correct all erroneous templates until all researchers agree that all templates are correct. Finally, we check for grammatical issues when filling subjects into templates, such as inconsistent pronouns. We find that the filled templates are reasonable in most cases, with an accuracy rate of 97.5\%.

After completing the above steps, we have finished the entire process of data collection and validation, which means we have completed the construction process of the dataset.

\subsection{\methodname~ Details}
\label{sec:model_datail}
We introduced a novel caching mechanism and subject extraction to \methodname~. Inspired by computer cache systems, the caching mechanism utilized by \methodname~ ensures that the stored knowledge is up-to-date while facilitating fast retrieval. Subject extract techniques allow \methodname~ to retrieve stored knowledge more precisely than previous techniques. In this section, we present the details of \methodname. 
\subsubsection{Entity Extraction}
\label{sec:Entityextraction}
Entity extraction aims to extract key entities from the provided input. Previous research has extensively explored methods such as NER or entity-linking  
 \citep{wu2019scalable}. We use LLMs to assist us in completing this task. Results indicate that this subtask can easily attain an accuracy rate exceeding 97\% on \methodname. The accuracy statistics of entity extraction on \methodname~ are depicted in the table \ref{tab:Entity extraction} and the prompt is in the figure \ref{fig:prompt_entity_ex}.
\begin{table}[!ht]
\centering
\begin{tabular}{cc}
\hline
\textbf{Method} & \textbf{accuracy}\\
\hline
GPT-3.5-turbo & $98.1$ \\
Llama2 & $97.3$ \\
T5 & $99.8$ \\
\hline

\end{tabular}
\caption{Accuracy for entity extraction, when using GPT-3.5-turbo and Llama2, we employ few-shot. When using T5, we finetune on \benchmarkname~items for 5 epochs.}
\label{tab:Entity extraction}
\end{table}

\begin{figure*}[!t]
\centering
\begin{minipage}{\textwidth}
\begin{lstlisting}[]
Given a sentence, identify and extract the primary entity mentioned. Ensure that the entity extracted does not include any punctuation or special characters. Your response should consist of the entity's name or title, such as a person's name, place, or organization. If the sentence contains multiple entities, select the most prominent or elevant one. 

##few-shot
What is the inspiration behind the name of Seine-Maritime?
Seine-Maritime

Who is the cast member of Casino Royale?
Casino Royale

##total 8 few-shot 
\end{lstlisting}
    \caption{Prompt for entity extraction}
    \label{fig:prompt_entity_ex}
\end{minipage}
\end{figure*}

\subsubsection{Knowledge Base Retrieval}
The local knowledge base is stored in the form of a knowledge graph. When updating the local knowledge base, it can be automatically updated from the external database or manually injected with certain facts to reflect real-world changes. Such updates may require a considerable amount of time, but they can be done in parallel in arbitrary quantities and during idle times. Consequently, we did not explicitly evaluate the duration dedicated to this aspect.

\subsubsection{Knowledge Rank and Utilization}

Following previous works \citep{zhong2023MQuAKE,zheng2023can}, we rank the retrieved knowledge based on similarity to the input and select the top-k knowledge. In our experiments, we set $k=1$. We prompt the model to use the retrieved knowledge for updating its output, which is shown in figure \ref{fig:prompt_entity_kn}. To ensure that the model's output meets the task requirements, we added a task prompt before all prompts, as shown in Figure \ref{fig:prompt_entity_task}.

We utilized an off-the-shelf retrieval model \citep{izacard2021unsupervised} to identify and rank the fact triplets, which allows us to avoid the training process.

\begin{figure*}[!t]
\centering
\begin{minipage}{\textwidth}
\begin{lstlisting}[]
<task_prompt>

##few-shot
(Hiroshima Prefecture, head of government, Hidehiko Yuzaki)
Q: Who is the leader of the government in Hiroshima Prefecture?
A: Hidehiko Yuzaki.

(Naples, head of government, Gaetano Manfredi)
Q: Who is the current head of government for Naples?
A: Gaetano Manfredi.

##total 3 few-shot
<Retrieved triples>
Q: <query>
A: 
\end{lstlisting}
    \caption{Prompt for knowledge rank and utilization}
    \label{fig:prompt_entity_kn}
\end{minipage}
\end{figure*}

\begin{figure*}[!t]
\centering
\begin{minipage}{\textwidth}
\begin{lstlisting}[]
completion: "Complete the sentence with a phrase."
qa: "Answer the question with one phrase."
local: "Answer the question with one phrase."
fill: "Identify the content within the parentheses and provide the missing information."
choose: "Choose the best answer."
fc: "Determine the veracity of the provided statement. Clearly output 'True' if the statement is accurate and 'False' if it is not." 
\end{lstlisting}
    \caption{Prompt for task adaptation.}
    \label{fig:prompt_entity_task}
\end{minipage}
\end{figure*}

\subsection{Implementation Details of Baselines} 
\label{sec:Baselines}
For FT, MEMIT, and IKE, we use the framework provided by \citet{wang2023easyedit}\footnote{\url{https://github.com/zjunlp/EasyEdit}}. For Mello, we used the original implementation but modified the prompt to fit tasks. \footnote{\url{https://github.com/princeton-nlp/MQuAKE}}

\paragraph{FT}
Following previous works \citep{meng2022mass}, We apply Fine-Tuning (FT) to the given layer of the model. For GPT2-XL, we select layer 0, and for GPT-J and Llama2, we choose layer 21. 

\paragraph{MEMIT}
For GPT2-XL and GPT-J, we employ default hyperparameters. For Llama2, we update the parameters of layers $\{4, 5, 6, 7, 8\}$. Across all models, we calculate covariance statistics using 50,000 instances from Wikitext.

\paragraph{MeLLo}
The original method was designed for multi-hop questions. We redesign the prompt for each task while keeping the knowledge retrieval part unchanged.

\paragraph{IKE}
In the original paper, relevant facts were directly added to the prompt. To make a fair comparison, we removed this part and ensured that all facts were retrieved\footnote{The author of IKE's response to the issue: \url{https://github.com/Zce1112zslx/IKE/issues/3}}. Our retrieval settings remained consistent with the original paper.

\paragraph{Other Baselines}
SERAC \citep{mitchell2022memory} and EREN \citep{chen2024robust} are two strong baselines. However, SERAC requires a significant amount of time for retraining \citep{yao2023editing}, making it difficult to handle frequently updated requests. EREN is suitable for models that have undergone instruction fine-tuning, while \methodname~ and others focus on base models. Therefore, we did not include these two baselines in the comparison.

\subsection{Other Benchmarks}
\label{sec:other_benchmarks}
To comprehensively evaluate model editing methods, we tested these methods on triviaQA \citep{joshi2017triviaqa}, Natural Questions \citep{kwiatkowski2019natural}, FEVER \citep{Thorne18Fever} and VitaminC \citep{schuster2021get}. TriviaQA and Natural Questions are commonly employed to assess the capabilities of LLMs \citep{touvron2023Llama}. FEVER serves as a classic dataset for fact-checking, and VitaminC has been utilized in prior works to evaluate the effectiveness of model editing \citep{mitchell2022memory}.

\section{Additional Experiment}
\label{app:add_exp}
\subsection{Ablation Study}
We conduct ablation experiments targeting different databases, demonstrating the necessity of selecting structured databases. As shown in Table \ref{tab:performance_comparison}, when editing the Llama2 model, we found that the limitations in retrieval accuracy of embedding databases resulted in significant side effects, leading to poor model performance.

\begin{table}[h]
    \centering
    \begin{tabular}{cccc}
    \hline
    \textbf{Databases} & \textbf{\scorename↑} & \textbf{EM↑} & \textbf{DD↓} \\ 
    \hline
    Structured Database & $90.54$ & $90.54$ & $0.00$ \\ 
    Embedding Database & $69.89$ & $89.64$ & $19.75$ \\ 
    \hline
    \end{tabular}
    \caption{Performance comparison of databases}
    \label{tab:performance_comparison}
\end{table}

\begin{table*}[!ht]
    \centering
\begin{tabular}[\textwidth]{cc}
  \hline
  \textbf{Symbol} & \textbf{Example} \\
  \hline
  s & America \\
  r & head of government \\
  o & Joe Biden \\
  $(x,y)$ & \makecell[c]{(Who is the current head of government for America?, Joe Biden)}  \\
  $I(x,y)$ & \makecell[c]{(The head of government for America is \_\_, Joe Biden)}  \\
  $EX(x,y)$ & \makecell[c]{(Who is the spouse of the President of the United States?, Jill Biden)}  \\
  $O(x,y)$ & (What color is the Sky?, Blue) \\
  \hline
\end{tabular}
    \caption{
    \label{app:example_def}Example of definition. The examples for $I(x,y)$, $EX(x,y)$, and $O(x,y)$ represent elements in the set.
    }
\end{table*}
\section{Example of Definition}
\label{sec:exampleOfDefination}
\label{sec:ProblemsDefinitionDetails}
Table \ref{app:example_def} shows an example of the definition.

\section{Detailed Comparison between \benchmarkname~and MQuAKE-T}
 \label{app:detail_cmp_datasets}
 
Similar to MQuAKE-T \citep{zhong2023MQuAKE}, \benchmarkname~ consists of genuine knowledge rather than constructed false information. However, MQuAKE-T is designed for multi-hop questions and is characterized by fewer relations and a smaller dataset size compared to ours (size: 1.8k vs. 8k, number of relations: 6 vs. 86), making it challenging to use it to enhance model capabilities. Therefore, we are currently the only benchmark available that can augment these capabilities.

\begin{figure}[!ht]
    \centering
    \includegraphics[width=\linewidth]{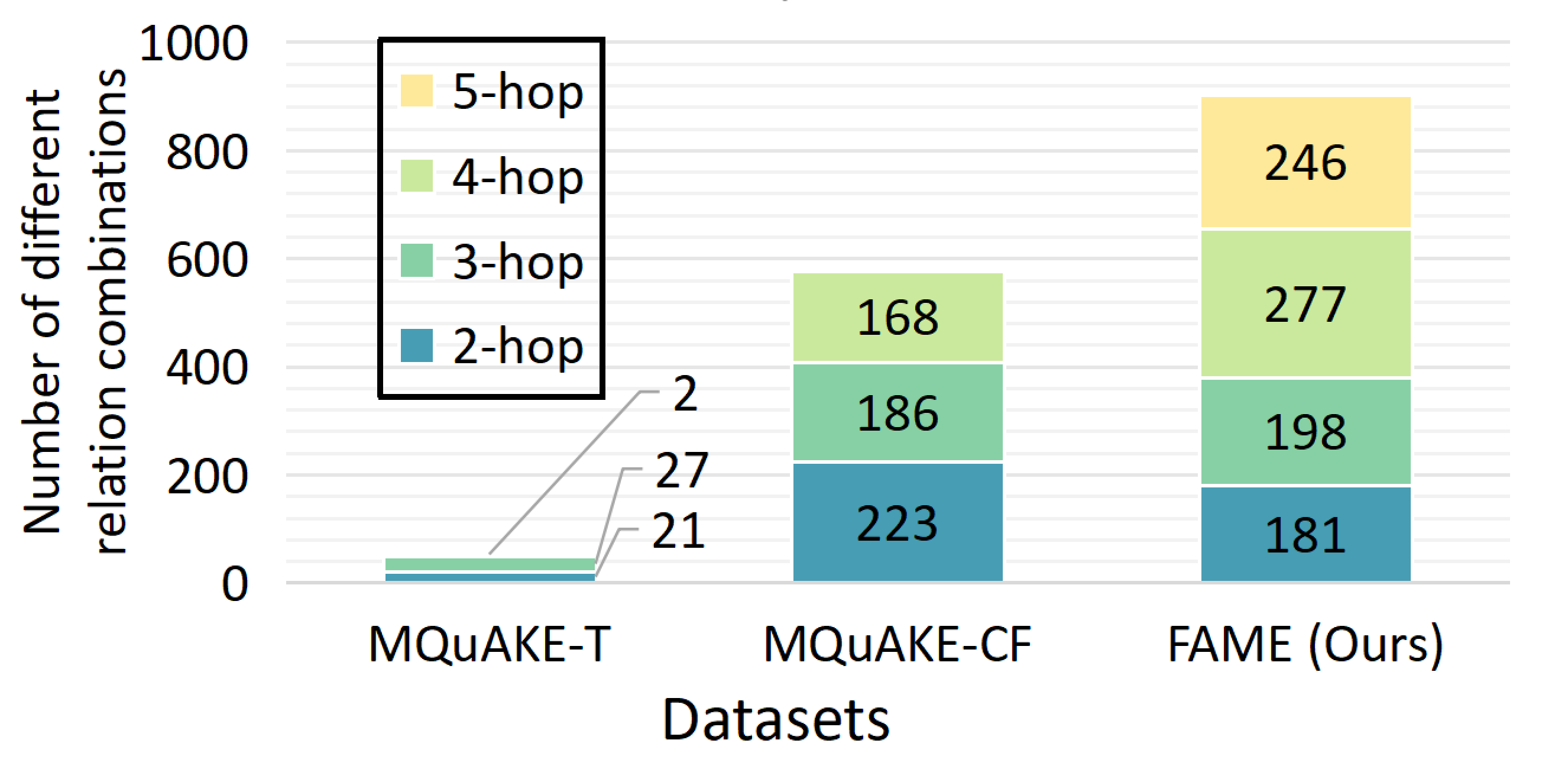}
    \caption{Comparison between multi-hop data in \benchmarkname~and MQuAKE. The vertical axis of the graph represents the number of relation combinations. \benchmarkname~encompasses a greater number of combinations, including 5-hop questions, which effectively demonstrates the enhanced diversity of our dataset.}
    \label{fig:cmp_multi_qa}
\end{figure}

    Figure \ref{fig:cmp_multi_qa} shows a comparison between relation combinations in our data and MQuAKE \citep{zhong2023MQuAKE}. It can be observed that our multi-hop questions cover a higher number of relationships, indicating that our data is more comprehensive.

\section{Examples and Statistics of \benchmarkname}
\label{app:example_sta_for_dataset}
\subsection{Examples of \benchmarkname}
Figure \ref{fig:example_single} shows an example of a single-hop question in \benchmarkname. The object label serves as the golden answer, with different queries used for corresponding tasks. Locality queries evaluate the model's output using similar (sharing the same relation) but unrelated queries. The locality object label represents the golden answer for the locality query.

\begin{figure*}[!ht]
\centering
\begin{minipage}{\textwidth}
\begin{lstlisting}[]
"subject_label": "Sioux Falls",
"relation_label": "head of government",
"object_label": "Paul Ten Haken",
"localitysubjectLabel": "Viarmes",
"localityobjectLabel": "William Rouyer"
"qa_query": "Who is the current head of government for Sioux Falls?",
"fill_query": "() is the head of government in Sioux Falls.",
"completion_query": "The head of government for Sioux Falls is",
"choose_query": "Who holds the position of head of government in Sioux Falls?\nA:Theodor Leutwein B:Lothar von Trotha C:Paul Ten Haken",
"FC_query": "Determine whether the proposition is true.\nProposition:The head of government for Sioux Falls is Paul Ten Haken."
"locality_query":"Who is the current head of government for Viarmes?"

\end{lstlisting}
    \caption{An example of single-hop questions in FAME}
    \label{fig:example_single}
\end{minipage}
\end{figure*}

\begin{figure*}[!ht]
\centering
\begin{minipage}{\textwidth}
\begin{lstlisting}[]
"s1_label": "Amtrak",
"relation_label_1": "owner of",        
"o1_label": "Route 128 station",
"relation_label_2": "located in the administrative territorial entity",
"o2_label": "Westwood",
"qa_query_1": "What entities does Amtrak owns?",
"qa_query_2": "In which administrative territorial entity is Route 128 station located?",
"MultihopQA_query":"In which administrative territorial entity is the entities {} owns located?"

\end{lstlisting}
    \caption{An example of multi-hop questions in FAME}
    \label{fig:example_multi}
\end{minipage}
\end{figure*}

For multi-hop questions, take the triple chain as an example: \textit{[(America, head of government, Biden), (Biden, spouse, Jill Biden)]}. Facing questions like \textit{"Who is the spouse of the head of government in America?"}, when Trump is updated to Biden, the answer to this question should change accordingly. In this example, "Biden" is the first object, "Jill Biden" is the second object, and the query is "Who is the spouse of the head of government in America?".
Figure \ref{fig:example_multi} shows an example of a 2-hop question in \benchmarkname.

\subsection{Statistics of \benchmarkname}

Table \ref{tab:relation_statistics} and Table \ref{tab:relation_statistics_c} display the labels of the relations we utilized in DBpedia and Wikidata,  along with the number of data items associated with each relation.

\begin{table*}[!ht]
    \centering
    \begin{tabular}{ccc}
    \hline
    \textbf{DBpedia Label} & \textbf{Wikidata Label} & \textbf{Relevant Triples} \\
    
    \hline
 prime minister & head of government & 199 \\ 
 highway system & transport network & 189 \\ 
 country & country & 298 \\ 
 birth place & place of birth & 251 \\ 
 death place & place of death & 246 \\ 
 sex & sex or gender & 351 \\ 
 father & father & 271 \\ 
 mother & mother & 243 \\ 
 spouse & spouse & 232 \\ 
 citizenship & country of citizenship & 226 \\ 
 capital & capital & 230 \\ 
 currency & currency & 217 \\ 
 child & child & 237 \\ 
 family & family & 216 \\ 
 team & member of sports team & 249 \\ 
 film director & director & 209 \\ 
 discoverer & discoverer or inventor & 262 \\ 
 alma mater & educated at & 238 \\ 
 architect & architect & 248 \\ 
 anthem & anthem & 255 \\ 
 composer & composer & 237 \\ 
 editor & editor & 223 \\ 
 discipline & field of work & 230 \\ 
 party & member of political party & 219 \\ 
 employer & employer & 220 \\ 
 illustrator & illustrator & 204 \\ 
 founded by & founded by & 218 \\ 
 league & league & 242 \\ 
 place of burial & place of burial & 240 \\ 
 maintained by & maintained by & 224 \\ 
 owner & owned by & 242 \\ 
 county & located in the administrative territorial entity & 218 \\ 
 movement & movement & 229 \\ 
 genre & genre & 223 \\ 
 named after & named after & 265 \\ 
 religion & religion or worldview & 263 \\ 
 based on & based on & 230 \\ 
 architectural style & architectural style & 225 \\ 
 headquarter & headquarters location & 239 \\ 
 starring & cast member & 251 \\ 
 chief executive officer & chief executive officer & 147 \\ 
 creator (agent) & creator & 259 \\ 
 builder & manufacturer & 223 \\ 
    \hline
    \end{tabular}
    \caption{Statistics of relations}
    \label{tab:relation_statistics}
\end{table*}

\begin{table*}[!ht]
    \centering
    \begin{tabular}{ccc}
    \hline
    \textbf{DBpedia Label} & \textbf{Wikidata Label} & \textbf{Relevant Triples} \\
    
    \hline
 crosses & crosses & 253 \\ 
 developer & developer & 237 \\ 
 doctoral advisor & doctoral advisor & 211 \\ 
 doctoral student & doctoral student & 248 \\ 
 construction material & made from material & 223 \\ 
 inflow & inflows & 258 \\ 
 IATA code & IATA airline designator & 225 \\ 
 ICAO code & ICAO airline designator & 233 \\ 
 record label & record label & 230 \\ 
 license & copyright license & 257 \\ 
 manager & head coach & 225 \\ 
 designer & designed by & 217 \\ 
 cinematography & director of photography & 253 \\ 
 is part of & part of & 237 \\ 
 original language & original language of film or TV show & 248 \\ 
 launch vehicle & space launch vehicle & 258 \\ 
 computing platform & platform & 253 \\ 
 Game Engine & software engine & 240 \\ 
 position & position played on team / speciality & 221 \\ 
 industry & industry & 246 \\ 
 colour & color & 253 \\ 
 homeport & shipping port & 252 \\ 
 general manager & general manager & 131 \\ 
 formation date & inception & 208 \\ 
 discovery date & time of discovery or invention & 244 \\ 
 start date & start time & 264 \\ 
 battle & conflict & 258 \\ 
 diocese & diocese & 224 \\ 
 cover artist & cover art by & 251 \\ 
 distributor & distributed by & 243 \\ 
 notable work & notable work & 233 \\ 
 crew member & crew member(s) & 239 \\ 
 editing & film editor & 228 \\ 
 fuel system & fuel system & 146 \\ 
 instrument & instrument & 234 \\ 
 eye color & eye color & 240 \\ 
 birth name & birth name & 225 \\ 
 owning organisation & owner of & 245 \\ 
 length ($\mu$) & length & 235 \\ 
 elevation ($\mu$) & elevation above sea level & 256 \\ 
 area total ($m^{2}$) & area & 230 \\ 
 runtime ($s$) & duration & 224 \\ 
 military branch & military unit & 231 \\ 
    \hline
    \end{tabular}
    \caption{Statistics of relations (Continued)}
    \label{tab:relation_statistics_c}
\end{table*}

\section{ Detailed Analysis of \scorename}
\label{app:detail_analysis_of_score}

Model editing aims to precisely correct the knowledge within an LLM without affecting unrelated knowledge. Therefore, measuring the accuracy of updates and the side effects is paramount. However, scenarios may have varying requirements for these two aspects. For example, in critical fields like medical diagnostics or legal advice, even minor inaccuracies can have severe repercussions, so we place more emphasis on the model's performance within the scope of knowledge. Conversely, for a well-performing model, the focus shifts toward correcting misconceptions while maintaining its existing capabilities. Thus, it is necessary to consider both the in-scope accuracy and out-of-scope side effects and adapt to the demands of different scenarios.

To address this challenge, we propose \scorename, a metric that comprehensively measures the effectiveness of model editing and estimates the capabilities of the edited model in real-world scenarios. \scorename~considers both accuracy and side effects and introduces hyperparameters \(\alpha\) and \(\beta\) to adjust the importance of accuracy and side effects. For instance, selecting \(\alpha > \beta\) signifies a higher priority for in-domain knowledge, which is advantageous in sectors like healthcare, law, or when editing models on private datasets. Conversely, setting \(\alpha < \beta\) suggests that minimizing side effects is more crucial, which is suitable for extensively used models already delivering online services. Thus, by adjusting the hyperparameters, \benchmarkname~can evaluate the practical capabilities of the model in different usage scenarios.

\end{document}